\documentclass{article}

\usepackage[final, nonatbib]{neurips_bdl2019}

\usepackage[style=numeric, backend=biber, maxcitenames=2, maxnames=2,maxbibnames=6, giveninits=true, uniquename=false,uniquelist=false,
			sortcites=true,doi=false,url=false, bibencoding=utf8]{biblatex} 
\DeclareLanguageMapping{english}{english-apa}
\renewbibmacro{in:}{}
\usepackage[utf8]{inputenc} 
\usepackage[T1]{fontenc}    
\usepackage{hyperref}       
\usepackage{url}            
\usepackage{booktabs}       
\usepackage{amsfonts}       
\usepackage{nicefrac}       
\usepackage{microtype}      
\usepackage{amsmath}        

\usepackage{tikz}
\usetikzlibrary[topaths]
\newcount\mycount

\title{Implicit Priors for Knowledge Sharing in Bayesian Neural Networks}

\author{
Jack K Fitzsimons\thanks{equal contribution} \hspace{15mm} Sebastian M Schmon\footnotemark[1] \hspace{15mm} Stephen J Roberts \\
 University of Oxford \\
 Oxford, United Kingdom 
}

\begin{document}

\maketitle

\section{Introduction}

Bayesian interpretations of neural network have a long history, dating back to early work in the 1990's \autocite{mackay1992bayesian, neal2012bayesian} and have recently regained attention \autocite[e.g.][]{blundell2015weight, gal2016uncertainty} because of their desirable properties like uncertainty estimation, model robustness and regularisation.

In this paper we concider the application of Bayesian models to knowledge sharing between neural networks. Knowledge sharing comes in different facets, such as transfer learning, model distillation and shared embeddings. All of these tasks have in common that learned "features" ought to be shared across different networks. 

Bayesian approaches offer a robust statistical framework to introduce prior knowledge into learning procedures. However, the tasks introduced above can be challenging in practice since "information" gained by one network such as learned features can be difficult to encode into prior distributions over networks that do not share the same architecture or even the same output dimension. We introduce here a Bayesian viewpoint that centres around features and describe a set of prior distributions derived from the theory of Gaussian processes and deep kernel learning that facilitate a variety of deep learning tasks in a unified way. 
In particular, we will show that our approach is applicable to knowledge distillation, transfer learning and combining experts.

\section{Bayesian Neural Networks}

In the standard Bayesian interpretation, neural network weights, denoted $w$, are random variables endowed with a prior distribution $p(w)$. Let us denote the dataset $\mathcal{D}_n = \{ (x_i, y_i); i=1,\ldots,n\}$ consisting of independent observations $(x_i, y_i)$, where $x_i \in \mathbb{R}^m$ is the input data and $y_i$ the labels or output. The object of interest is the posterior distribution
\begin{equation*}
    p(w \mid \mathcal{D}_n) \propto \prod_{i=1}^n p(x_i, y_i \mid w) p(w) = \mathcal{L}(w; \mathcal{D}_n)p(w),
\end{equation*}
where we write $\mathcal{L}(w; \mathcal{D}_n)$ for the likelihood function.
Many tasks in deep learning naturally lend themselves to a Bayesian approach where a teacher network provides prior knowledge which is incorporated in the learning process of a student network.
However, assigning a prior on the weights directly is impractical as we are often interested in sharing information between networks with different architectures.
For this reason, we propose to distil \emph{features}, denoted $\phi$, generated by the teacher network, i.e.
\begin{equation*}
    p(\phi\mid \mathcal{D}_n) \propto \mathcal{L}(\phi ; \mathcal{D}_n)p(\phi),
\end{equation*}
where $p(\phi)$ is the prior for the features. For prior elicitation we draw from the theory of Gaussian processes which arise naturally in the context of neural networks although other approaches are possible. For a detailed introduction to Gaussian processes we refer the reader to \autocite{rasmussen2003gaussian}. In the following we will briefly review how feature spaces created by neural networks form Gaussian processes.

\subsection{Neural Networks as Gaussian Processes}
\label{sec:gp}

The Gaussian process interpretation of neural networks originates from early work by \autocite{neal2012bayesian} where it is shown that an infinite width single layer neural network is equivalent to a Gaussian process. This line of work has recently found renewed attention \autocite{matthews2018gaussian, lee2018deep}, where the authors consider deep networks. Let $x\in \mathbb{R}$ be an input and for layers $l=1, \ldots, L$ with width $N_l$ write
\begin{equation*}
    \phi_i^l(x) = b_i^l + \sum_{j=1}^{N_l}W_{ij}^l g(\phi_j^{l-1}(x)),
\end{equation*}
where $b_i^l$ denotes the bias of feature $i$ in layer $l$ and the weights $W_{ij}^l$ analogously. Here $g$ denotes some activation function. An application of a central limit theorem under Gaussian weights $W_{ij}^l$ shows that $N_l\rightarrow \infty$ for all layers induces a zero mean Gaussian process over features $\phi$ with covariance function
\begin{equation*}
    c_{ij}(x, x') = \delta_{i,j}\mathbb{E}\left[\phi_i(x) \phi_i(x') \right] \quad \text{for all } i, j.
\end{equation*}
For finite layer neural networks, we take \begin{equation*}
    \left(K_\phi(x, x')\right)_{i,j} = \phi_i^L(x)\phi_j^L(x').
\end{equation*}
This is can also be viewed as an instance of a deep kernel GP, see e.g. \autocite{wilson2016deep}, although we consider degenerate linear kernels (also referred to as dot product kernels) in this inferred feature space rather than RBF or spectral kernels.

\subsection{Distance Priors and Kullback-Leibler Divergence}

In order to pass the knowledge of a teacher network to a student, a good prior distribution places high probability on features that are similar to features of the teacher network. Denote $\hat\phi$ the features generated by the teacher network. Following our argument, a natural choice for a prior is then based on the distance between features
\begin{equation*}
    \log p(\phi) = - \alpha \cdot d(\phi, \hat\phi) + \text{constant},    
\end{equation*}
where $d$ is a some non-negative function measuring similarity of features but not necessarily a metric and $\alpha > 0$ is a tuning parameter. 

Approaches comparing features directly have been proposed in the past. Consider, for example, the case of model distillation \autocite{bucilua2006model, hinton2015distilling}, covered in more detail in the next section. If we choose independent priors for the features $\phi_i$ of the student network and $d$ is taken as 
\begin{equation*}
    d(\phi_i, \hat\phi_i) = \mathcal{H}\big(\sigma(\phi/T)_i, \sigma(\hat\phi/T)_i  \big),
\end{equation*}
for some temperature $T$, binary cross-entropy $\mathcal{H}$, and $\sigma$ the softmax function, we recover the approach by Hinton \autocite{hinton2015distilling}. However, such approaches have limitations. As well as to the unrealistic independence assumption, the above approach requires as many logits in the student model as we have in the teacher model. Similarly, we can not easily share information of previous layers. We circumvent this by comparing the distributions of the features on their induced function spaces using the KL divergence. The new prior distribution for the student network now reads
\begin{equation*}
    \log p(\phi) = - \alpha D_{\textsc{kl}}(\nu_\phi\,\|\,\nu_{\hat{\phi}}) + \text{constant}.
\end{equation*}
Since the feature maps are Gaussian processes, the KL divergence has an analytic form,
\begin{align*}
    & D_{\textsc{kl}}\big(\mathcal{GP}(\mu_1,K_1)\,\|\, \mathcal{GP}(\mu_2,K_2)\big) \\  &= \frac 1 2 \Big(\textnormal{Tr}(K_2^{-1}K_1) + (\mu_2\!-\!\mu_1)^\top K_2^{-1}(\mu_2\!-\!\mu_1)-n+\log\frac{|K_2|}{|K_1|}\Big).
\end{align*}
In doing so the KL divergence places a probability distribution over the space of features using the Gaussian processes, $\mathcal{GP}(0,K_\phi)$, parameterised by kernel $K_\phi$.
Note that this alleviates the requirement that the dimensionalities of the feature space (i.e. the number of neurons in the final layer) of the teacher and student network have to match as both are seen as functions in the same Hilbert space.
The prior is then
\begin{align*}
    p(\phi) & \propto \exp\left(- D_{\textsc{kl}}\big(\mathcal{GP}(0,K_{\phi}) \| \mathcal{GP}(0,K_{\hat\phi})\big)\right) \\ 
    & = \frac{|K_{\hat\phi}|}{|K_{\phi}|}\exp\left(-\frac{1}{2} \textnormal{Tr}(K_{\hat\phi}^{-1}K_\phi)\right).
\end{align*}
Other alternative choices for $d$ include Wasserstein distances, Hellinger distance, $L^2$-distance (between the features) and many more. However, we found that choices other than the KL-divergence did not improve our results. It is also worth noting that using KL-divergences in this way has clear links to the popular sparse variational Gaussian process and other approximate methods \autocite{hensman2015scalable}.

\section{Bayesian Knowledge Transfer}
\subsection{Model Distillation}
\label{sec:dist}

Using our approach, the concept of model distillation merely becomes a Bayesian neural network where the prior is (for example) our KL-prior derived from the features learned by the teacher model. As already alluded to earlier, these priors describe the behaviour of the features in more detail than a simple comparison of logits or a binary cross-entropy function.
In addition, unlike traditional model distillation the latent Hilbert space representation of the the model is not constrained by the dimensionality of the output logits of the teacher model. 

\subsection{Transfer Learning}
Transferring learned features from one model to another can be achieved similarly to the case of model distillation. 
It is important to note that the term "features" is not limited to the final layer logits. For example, in a convolutional neural network we could transfer the features learned by the convolutional layers disregarding the fully connected layers.

\subsection{Combining Experts}

Suppose we have a set of $m$ tasks $\tau_j, j=1, \ldots, m$ associated with a neural network that has learned its respective task. We want to combine the knowledge of those "experts" into one model. In order to do, we use the respective features $\hat{\phi}_j, j=1, \ldots, m$ and combine them as independent priors
\begin{equation*}
    p(\phi \mid \mathcal{D}_n) \propto \prod_{i=1}^n p(x_i, y_i \mid \phi_1, \ldots, \phi_m) p(\phi_1 \mid \hat{\phi}_1)\cdot \ldots \cdot p(\phi_m \mid \hat{\phi}_m).
\end{equation*}

\section{Example Application: Fully Connected Networks for Fashion-MNIST}

The first example application examines the benefit of using feature priors as a form of model distillation. The dataset considered is Fashion-MNIST. A classic convolutional neural network with two convolutional layers is composed of $3\times3$ filters and a dense layer with 128 nodes achieves an accuracy of 92.7\%. The goal of this exercise is to endeavour to train a fully dense network with two hidden layers. Intuition would suggest that a dense network naively will not perform very well at this task. However, by placing a prior on the output of the first dense layer to match the features learned by the teacher network we see an improvement by over 7\% in absolute terms. Both networks received 25 epochs and had the same architecture, loss function (excluding the prior component) and optimiser.

\begin{table*}[h!]\centering
\begin{tabular}{@{}rrrcrrcrr@{}}\toprule
& \multicolumn{2}{c}{Accuracy} & \phantom{abc}& \multicolumn{2}{c}{$F_1$-score (Micro)} &
\phantom{abc} & \multicolumn{2}{c}{$F_1$-score (Macro)}\\
\cmidrule{2-3} \cmidrule{5-6} \cmidrule{8-9}
& avg. & std. error && avg. & std. error && avg. & std. error \\ \midrule
Naive & 80.65\% & 1.02 && 80.65\% & 1.02 && 77.45\% & 1.47 \\
\autocite{ba2014deep} & 83.51\% & 0.14 && 83.51\% & 0.13 && 83.47\% & 0.15 \\
\autocite{hinton2015distilling} & 86.61\% & 0.11 && 86.62\% & 0.11 && 86.60\% & 0.11 \\
\autocite{sau2016deep} & 82.69\% &  0.12&& 82.68\% & 0.12 && 82.62\% & 0.14 \\
\autocite{romero2014fitnets} & 88.51\% & 0.07 && 88.52\% & 0.07 && 88.55\% & 0.08 \\
\emph{Proposed Approach} & \textbf{89.81\%} & 0.05 && \textbf{89.81\%}& 0.05&& \textbf{89.81\%}& 0.04\\
\bottomrule
\end{tabular}
\caption{Performance comparison of model distillation approaches on Fashion-MNIST. The models were both trained 30 times with random initialisation and the average and expected error on each metric was reported. }
\label{tab1}
\end{table*}

An important technical point to mention is how we dealt with the unknown tuning parameter $\alpha$ outlined in section 2.2. To avoid manual tuning, we first maximised the likelihood of the features agnostic of the training labels and then trained only remaining dense layers given the inferred features. Hence, our training was broken into two parts; training the features to match the teacher network followed by training the remaining layers to maximise the predictive performance. 

\begin{figure}[h!]
  \centering
  \includegraphics[width=0.6\textwidth]{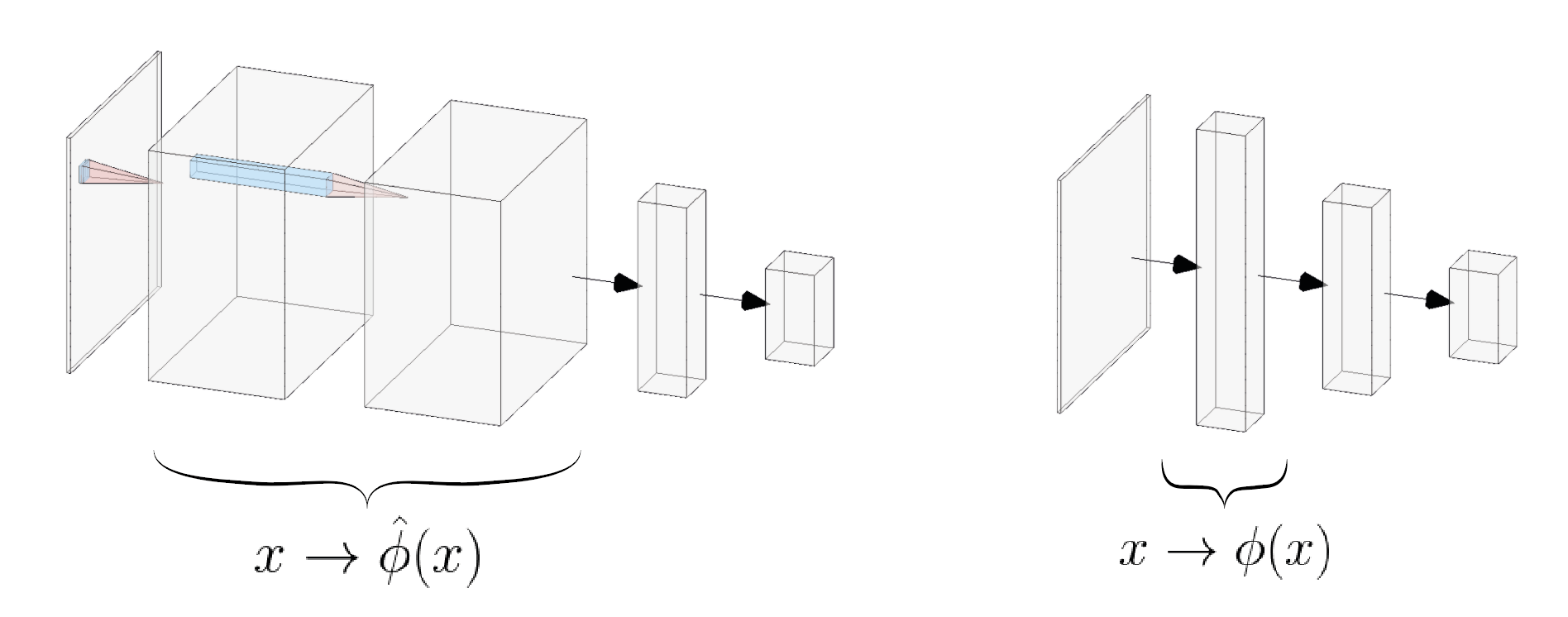}
  \caption{A comparison of the parent network and student network. On the left is the parent network composing of two convolutions layers followed by a dense layer and finally a softmax classification. The student model on the right endeavours to compress the two convolutional layers into a single dense layer.}
\end{figure}

\section{Example Application: Multi-Level Priors for CIFAR-10}

The second example application also inspects model distillation but compares the proposed approach to that set out in Hinton et al. \autocite{hinton2015distilling}. We purposefully chose a more complex network to demonstrate the benefits of the proposed approach. The teacher network composes of 4 VGG-like convolutional layers as depicted in Figure 2. As compressing multiple convolutional layers into a single dense layer would be seemingly more difficult, we split the convolutional feature extraction into two parts; the first two corresponding to low level features and the latter two corresponding to higher level features. A dense layer of 8192 hidden node was used to infer each of these sets of features. Comparing accuracy to the use of no parent model and even that of earlier approaches, which both did not significantly outperform a random classifier, shows the unparalleled benefit of such Bayesian knowledge transfer. 

\begin{table*}[h!]\centering
\begin{tabular}{@{}rrcrcrcr@{}}\toprule
& No Parent & & Classic Distillation &
 & Proposed Approach &  & Parent Model\\
\midrule
Top-1 Accuracy & 10.00\% && 10.00\% && \textbf{51.90}\% && \emph{65.60}\% \\
Top-2 Accuracy & 19.98\%&& 19.95\%&& \textbf{71.64}\% && \emph{82.08}\% \\
Top-3 Accuracy & 30.03\%&&  30.68\%&& \textbf{81.75}\% && \emph{89.77}\% \\
\bottomrule\end{tabular}
\caption{The above table compares the accuracy of the CIFAR10 for the naive dense network (no parent), utilising the method in \autocite{hinton2015distilling} and using the proposed approach.}
\end{table*}

Finally we note that this layer-wise distillation across architectures is not possible with \autocite{hinton2015distilling} as the number of outputs per intermediate layer do not generally match for different network architectures.

\begin{figure}[h!]
  \centering
  \includegraphics[width=\textwidth]{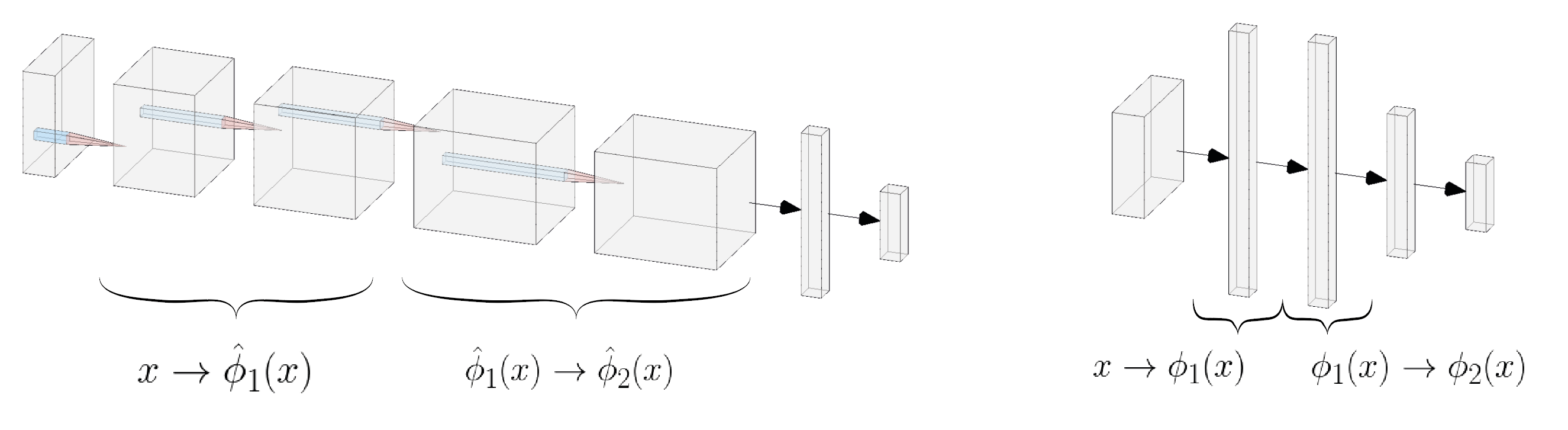}
  \caption{A comparison of the parent network and student network used for CIFAR-10. Similar to Figure 1, the figure shows a parent network composing of a CNN with 4 convolutional layers (left) and a fully dense student network (right). The convolutional layers are grouped into 2 and each is used in training the respective first two layers of the fully dense student network.}
\end{figure}

\paragraph*{Acknowledgements}
Sebastian M. Schmon’s research is supported by the Engineering and Physical Sciences Research Council (EPSRC) grant EP/K503113/1.

\printbibliography

\end{document}